\documentclass[runningheads]{llncs}
\usepackage{graphicx}
\usepackage{algorithm}
\usepackage{amsmath}
\usepackage{booktabs}
\usepackage[switch]{lineno}
\usepackage{amssymb}
\usepackage{bm}
\usepackage{subfigure}
\usepackage{multirow}
\usepackage{orcidlink}
\usepackage{algpseudocode}
\usepackage[misc]{ifsym}

\newcommand*\tcircle[1]{%
  \raisebox{-0.5pt}{%
    \textcircled{\fontsize{7pt}{0}\fontfamily{phv}\selectfont #1}%
  }%
}

\begin{document}
\title{CrimeAlarm: Towards Intensive Intent Dynamics in Fine-grained Crime Prediction}
\titlerunning{CrimeAlarm}
\author{Kaixi Hu\inst{1,2}\orcidlink{0000-0002-6774-8510} \and
Lin Li\inst{1}\Letter\orcidlink{0000-0001-7553-6916} \and
Qing Xie \inst{1}\orcidlink{0000-0003-4530-588X} \and
Xiaohui Tao \inst{3}\orcidlink{0000-0002-0020-077X} \and
Guandong Xu \inst{2,4}\orcidlink{0000-0003-4493-6663}
}

\authorrunning{K. Hu et al.}
\institute{Wuhan University of Technology, Wuhan, China.\\
\email{\{issac\_hkx, cathylilin, felixxq\}@whut.edu.cn} \\ \and
University of Technology Sydney, Sydney, Australia. \email{Guandong.Xu@uts.edu.au}\\  \and
University of Southern Queensland, Toowoomba, Australia.\\ \email{Xiaohui.Tao@unisq.edu.au}  \\  \and
Education University of Hong Kong, Hong Kong, China. \email{gdxu@eduhk.hk}}
\maketitle              % typeset the header of the contribution
\begin{abstract}
Granularity and accuracy are two crucial factors for crime event prediction. Within fine-grained event classification, multiple criminal intents may alternately exhibit in preceding sequential events, and progress differently in next. Such intensive intent dynamics makes training models hard to capture unobserved intents, and thus leads to sub-optimal generalization performance, especially in the intertwining of numerous potential events. To capture comprehensive criminal intents, this paper proposes a fine-grained sequential crime prediction framework, CrimeAlarm, that equips with a novel mutual distillation strategy inspired by curriculum learning. During the early training phase, spot-shared criminal intents are captured through high-confidence sequence samples. In the later phase, spot-specific intents are gradually learned by increasing the contribution of low-confidence sequences. Meanwhile, the output probability distributions are reciprocally learned between prediction networks to model unobserved criminal intents. Extensive experiments show that CrimeAlarm outperforms state-of-the-art methods in terms of NDCG@5, with improvements of 4.51\% for the NYC16 and 7.73\% for the CHI18 in accuracy measures.

\end{abstract}

\section{Introduction}
Crime can be thought to incorporate various activities that are deemed harmful to individuals, communities, or society. Crime prediction is an effective approach to prevent crime activities.
Although some ethical implications~\cite{hu2023information} are concerned, existing methods still show promising performance to make preliminary screening for possible crime events and assist humans in patrol allocation. Traditional crime prediction methods can be traced from crime mapping~\cite{levine2006crime}, which assesses the density, location, stability, and significance of historical crime events to determine crime hotspots. Furthermore, there have been user interaction softwares developed to leverage these methods, aiming to enhance criminal justice researches with crime mapping (e.g. CrimeStat). 

\begin{figure}[t]
\centering
  \includegraphics[width=0.85\linewidth]{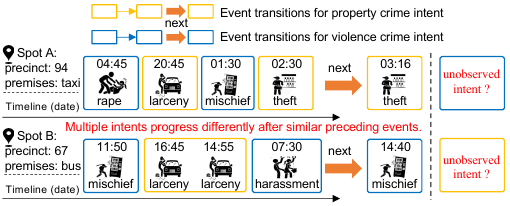}
  \caption{An illustration of intent dynamics behind crime events. Criminal intents are grouped into either property or violence. It is observed that both spot A and spot B contain multiple criminal intents (as flagged by different colors), and these intents switch frequently and/or progress differently.}
  \label{fig1}
 \vspace{-10pt}
\end{figure}      

Recent crime prediction studies have progressed to deep learning. Many efforts~\cite{hu2021duronet,DBLP:journals/inffus/HuLTVD23,DBLP:conf/cikm/HuL0LT21,DBLP:conf/icde/LiHXXP22,DBLP:conf/dasfaa/LinZW0L021} focus on \emph{sequential crime prediction} that exploits temporal dependencies behind historical event transitions to discover future criminal intents. Conceptually, a specific crime event is described by three elements: time (when), spot (where) and category (what happened)~\cite{DBLP:journals/ws/HageMSHS11,scherp2009f}. Fine-grained prediction may expect the partition of these elements are as small as possible. As such, existing methods can be divided into two groups: (1) The prediction of crime amounts~\cite{hu2021duronet,DBLP:conf/icde/LiHXXP22} is not rigidly \emph{fine-grained}, since the categorical information is eliminated by the amount of crime events in statistics; (2) The prediction of next specific crime category of each spot is a challenging task. Most works~\cite{DBLP:conf/www/HuangZZWCY19,DBLP:conf/cikm/HuangZZC18,DBLP:conf/aaai/WangLYSYS22} attempted to predict binary crime occurrences of all crime categories simultaneously. Recent HAIL~\cite{DBLP:conf/cikm/HuL0LT21} further predicted the ranking of different crime categories, which is more fine-grained.

\emph{Fine-grained Crime Prediction} incurs \emph{Intensive Intent Dynamics.} From criminology studies~\cite{hu2023information,wikstrom2014crime}, we can learn that regional criminal activities are dynamic, characterized by diverse intents that constantly vary according to real-time environments. In particular, the observed intent dynamics may be amplified in fine-grained event classification. Fig.~\ref{fig1} shows that criminal intents, as flagged by different colors, may alternately exhibit in preceding sequential events, occurring in spot A or spot B over time. The frequent switching among multiple criminal intents leads to different next observations, which reflects \emph{intensive intent dynamics}. This makes deep prediction networks hard to model unobserved intents, especially in the intertwining of numerous candidate events. Properly modeling the intensive intent dynamics is crucial to capture unobserved intents and boost generalization performance.

To this end, this paper proposes CrimeAlarm, a fine-grained crime prediction framework with a novel distillation strategy. It employs distinctive probability distributions (soft target) from different prediction networks to reciprocally learn unobserved intents, and finally derive a comprehensive intent in themselves intent representation spaces. In particular, fine-grained crime event division may lead to numerous candidate crime events. To enhance intent learning among these events, a novel distillation strategy is devised according to our theoretical analysis. This process starts from spot-shared intents and then gradually turns to spot-specific intents, following the easy-to-hard cognition~\cite{DBLP:conf/icml/BengioLCW09}.

\paragraph{Contributions.} Our contributions are as follows: (1) The intensive intent dynamics is highlighted as a challenge of fine-grained crime prediction. (2) We propose a novel fine-grained crime prediction framework (CrimeAlarm) according to our theoretical analysis. (3) Experiments are conducted on two real-world crime datasets to validate the effectiveness of CrimeAlarm.

\section{Preliminaries}
\subsection{Problem Statement}
Let $\mathcal{O}$ denote a spot set describing where a crime occurred and $\mathcal{I}$ denote a crime event class set describing what happened. For each spot $o \in \mathcal{O}$, the crime events where spot $o$ occurs are ordered chronologically, and obtain an event sequence $[\text{event}_1,...,\text{event}_l]$, where $\text{event}_l\in\mathcal{I}$ is the crime event class that happened at time step $l$. This work mainly focuses on the modeling of temporal information inside sequences. The spatial information is implicitly modeled in the spot set, and we will discuss more effective spatial modeling in future work.

\paragraph{Fined-grained Sequential Crime Prediction.} Given a training set $\mathcal{D} = (\mathcal{X},\mathcal{Y})$, $\bm{x}_m$ represents the $m$th sequence comprising $L-1$ crime event classes occurred in the past, and $\bm{y}_m$ denotes the target event class at time step $L$. $L$ denotes the maximum sequence length. This work aims to develop a network model $\mathcal{F}: \mathcal{X} \rightarrow \mathcal{Y}$ that is capable of inferring the probability distribution $\bm{p} =[p_{1},...,p_{\left|\mathcal{I}\right|}]$ of candidate crime event classes, where 
\begin{small}
\begin{align}
p_{i}=\frac{\text{exp}(z_i)}{\sum_{j=1}^{\left|\mathcal{I}\right|}\text{exp}(z_j)},
\label{sofmax}
\end{align}
\end{small}%
and $z_i$ is the logit of the $i$th event. Finally, the event with the maximum probability is the predicted next crime event $\hat{\bm{y}}$.

%that larger number of classes will increase the difficulty of distillation and suppress the learning of probability distribution. the target part of vanilla KD is related the training difficulty of samples, and the non-target part provides knowledge among classes. This knowledge shows the relative relation among non-target classes to present in the next, which is crucial for modeling dynamics.
 
\subsection{Decoupled Knowledge Distillation}
\paragraph{Why knowledge distillation works?} Different from one-hot labels (hard target), the probability distributions of event classes (soft target) provide knowledge among unobserved intents (a.k.a. dark knowledge~\cite{DBLP:conf/icml/FurlanelloLTIA18}). Knowledge distillation (KD) is an effective paradigm to transfer such knowledge from teacher networks to student networks and obtain better generalization performance. 

Recent decoupled knowledge distillation (DKD)~\cite{Zhao_2022_CVPR} reveals that the target distillation is related to training difficulty, and the non-target distillation provides knowledge among classes. For the $k$th prediction network, given its logits $\bm{z}^{(k)}$, the cross-entropy (CE) of DKD can be written as:
\begin{small}
\begin{align}
\begin{split}
\mathcal{L}_{\text{DKD}}^{(k)}&=\text{CE}({\tilde{\bm{p}}_{_\text{TC}}^{(k)}}\|{\bm{p}_{_\text{TC}}^{(k)}})+\beta \cdot \text{CE}({\tilde{\bm{q}}_{_\text{NC}}^{(k)}}\|{\bm{q}_{_\text{NC}}^{(k)}})\\
&=\underbrace{\underbrace{-\tilde{p}_*^{(k)}\text{log}(p_*^{(k)})}_{\tcircle{1}} - \underbrace{\tilde{p}_{\backslash*}^{(k)}\text{log}(p_{\backslash*}^{(k)})}_{\tcircle{2}}
}_{\textbf{Target}}-\beta\cdot \underbrace{\sum_{i=1,i\neq*}^{\left|\mathcal{I}\right|}\tilde{q}_i^{(k)}\text{log}(q_i^{(k)})}_{\substack{
\tcircle{3}\; \textbf{Non-target}}},
\end{split}
\label{kd_decouple}
\end{align}
\end{small}%
where $\beta$ is a hyper-parameter of non-target distillation, and the target probability distribution of a student model is
\begin{small}
\begin{align}
\bm{p}_{_\text{TC}}^{(k)}=\left[p_*^{(k)},\; p_{\backslash*}^{(k)}\right]=\left[\frac{\text{exp}(z_*^{(k)})}{\sum_{j=1}^{\left|\mathcal{I}\right|}\text{exp}(z_j^{(k)})},\; \frac{\sum_{l=1,l\neq*}^{\left|\mathcal{I}\right|}\text{exp}(z_l^{(k)})}{\sum_{j=1}^{\left|\mathcal{I}\right|}\text{exp}(z_j^{(k)})}\right],
\label{p_ti}
\end{align}
\end{small}%
while the non-target probability distribution of a student is
\begin{small}
\begin{align}
\begin{split}
\bm{q}_{_\text{NC}}^{(k)}&=[q_1^{(k)},...,q_{i}^{(k)},...,q_{\left|\mathcal{I}\right|}^{(k)}]\in\mathbb{R}^{1\times {(\left|\mathcal{I}\right|-1)}},\\
q_{i}^{(k)}&=\frac{\text{exp}(z_i^{(k)})}{\sum_{j=1,j\neq *}^{\left|\mathcal{I}\right|}\text{exp}(z_j^{(k)})}, \quad \text{s.t.} \; \sum_{i=1,i\neq*}^{\left|\mathcal{I}\right|}q_i^{(k)}=1
\end{split}
\label{q_ni}
\end{align}%
\end{small}%
and ``$*$'' is the index of the target (ground-truth). For simplicity, we use tilde ($\sim$) as the symbol of corresponding probability distribution from its teacher.

Basically, each sequence sample is involved with a target crime event (observed intent) while other unobserved intents are covered in numerous non-target crime events. It is appropriate to model them separately through DKD. However, based on our follow-up analysis in Sec.~\ref{sec32} and Sec.~\ref{sec33} about target and non-target parts in Eq.~(\ref{kd_decouple}), we find it is intractable for existing KD to handle numerous candidate crime events. This further inspire our CrimeAlarm.

%which is appropriate for a modeling task through decoupled KD.

\section{CrimeAlarm}
\label{crimewhere}
\subsection{Overview}
Our CrimeAlarm aims to model the intensive intent dynamics by concerning on the output logits during training process. This makes CrimeAlarm a model-agnostic distillation framework that integrates various general or specific-domain sequential prediction networks without respect to their structure difference. And, due to the lack of powerful pre-trained teacher network in crime prediction, this work follows online mutual distillation~\cite{DBLP:conf/eccv/QianWH,DBLP:conf/cvpr/ZhangXHL18} where dual-peer prediction networks are reciprocally trained from scratch and improved simultaneously by exploiting their response difference.

\begin{figure}
  \centering
  \includegraphics[width=0.8\linewidth]{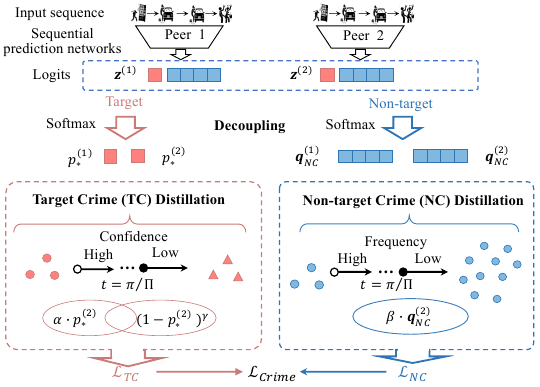}
  \caption{The framework of our proposed CrimeAlarm. In the early distillation phase, spot-shared intents are learned via simple target, and less yet frequent non-target crime events. In the later phase, spot-specific intents are captured via difficult target and more non-target crime events.}
  \label{fig:framework}
\end{figure}

As shown in the upper part of Fig.~\ref{fig:framework}, with the logits derived by peer networks, the decoupled probability distributions are calculated based on DKD. In this work, to enhance distillation among numerous candidate crime events, we further introduce training progress $t=\pi/\Pi$ into Eq.~(\ref{kd_decouple}), where $\pi$ is the current training epoch and $\Pi$ is the total epochs. The easy-to-hard pipeline is inspired by curriculum learning and can be described as follows:
\begin{small}
\begin{align}
    \mathcal{L}_\text{DKD}^{(k)} \Rightarrow \mathcal{L}_{\text{Crime}}^{(k)} 
    = \underbrace{\Gamma( t, \mathcal{L}_{_\text{sim}}^{(k)},\mathcal{L}_{_\text{diff}}^{(k)} )}_{\textbf{Target:\,}\mathcal{L}_{_\text{TC}}^{(k)}}
    +\beta \cdot \underbrace{\text{CE}({\tilde{\bm{q}}_{_\text{NC}}^{(k)}(t)}\|{\bm{q}_{_\text{NC}}^{(k)}(t)})}_{\textbf{Non-target:\,}\mathcal{L}_{_\text{NC}}^{(k)}}.
\end{align}
\end{small}%
The above formulation includes two parts: (1) For the \textbf{target crime distillation}, we devise a novel difficult training for the later phase ($\mathcal{L}_{_\text{diff}}^{(k)}$), while keeping the existing simple training ($\mathcal{L}_{_\text{sim}}^{(k)}$) in the early phase. $\Gamma(\cdot)$ is a function to switch the early and later training according to $t$. (2) For the \textbf{non-target crime distillation}, we employ $t$ to mask non-target crime events in their probability distribution $\bm{q}_{_\text{NC}}^{(k)}$.

\subsection{Simple-to-Difficult Target Crime Distillation}
\label{sec32}
\paragraph{Analysis about Target.}
In line with~\cite{DBLP:conf/icml/FurlanelloLTIA18,DBLP:journals/corr/RomeroBKCGB14}, KD increases the contribution of high-confidence sequence samples. This conclusion can also be derived from the target term $\tcircle{1}$ in Eq.~(\ref{kd_decouple}), where the probability $\tilde{p}_*^{(k)}$ is the importance weight of samples. The term $\tcircle{2}$ suppresses the term $\tcircle{1}$ on low confidence samples. In terms of observed criminal intents, we believe, spot-shared intents are a type of common phenomenon that are easy to learn through high-confidence samples. However, there also exist some low-confidence samples with spot-specific crime intents. Such samples are more informative to learn comprehensive criminal intents and it is not advised to overlook them. Inspired by curriculum learning~\cite{DBLP:conf/icml/BengioLCW09}, we devise novel difficult target crime distillation, that enhances spot-specific intents in the later training phase.

\paragraph{Difficult Target Crime Distillation.}
Given the probability $p_{*}^{(k)}$ of target event, the loss can be formally defined as:
\begin{small}
\begin{align}
\mathcal{L}_{\text{diff}}^{(k)}(\tilde{p}_{*}^{(k)}\Vert p_{*}^{(k)})=-(1-\tilde{p}_{*}^{(k)})^{\gamma} \text{log}(p_{*}^{(k)}),
\label{loss_late}
\end{align}
\end{small}%
where $\gamma$ is a hyper-parameter to tune the importance of difficulty $(1-\tilde{p}_{*}^{(k)})^{\gamma} $, and $\tilde{p}_{*}^{(k)}$ is the probability derived from its peer. To avoid fitting noisy samples~\cite{DBLP:conf/wsdm/WangF0NC21}, 
We further posit that a sample might exhibit noise if it fail to be accurately predicted by all networks~($\forall k$ network). Therefore, the probability distribution $\tilde{p}_{*}^{(k)}$ is truncated as:
\begin{small}
\begin{align}
  \tilde{p}_{*}^{(k)}=
\begin{cases}
    1,  & \text{rank}(p^{(\forall k)}_{*}) < \varepsilon \cdot {\vert\mathcal{B}\vert}\\
    \tilde{p}_{*}^{(k)},  & \text{otherwise}
\end{cases}
\label{denoise}
\end{align}
\end{small}%
where $\varepsilon$ is the truncation proportion and $\vert\mathcal{B}\vert$ is the batch size of training samples and rank($\cdot$) denotes the probability rank among samples in ascending order.

For difficult target crime distillation, we note three properties as follows:
\begin{enumerate}
  \item For truncated sequence samples, $\tilde{p}_{*}^{(k)}$ is reset to 1 and the importance of difficulty goes to 0, thus the $k$th network will reduce the learning of it;
  \item For samples consistently yielding correct (or incorrect) responses, the probabilities $\tilde{p}^{(k)}_{*}$ and $p^{(k)}_{*}$ tend to be comparatively higher (or lower) across the dataset. Consequently, the significance of difficulty diminishes (or intensifies), leading to reduced (or increased) loss values.
  \item For samples displaying inconsistent responses, the significance of difficulty lies between the two cases described in property (2). Consequently, these samples can be categorized into three levels for training.
\end{enumerate}

\paragraph{Target Crime Distillation Loss.}
To integrate the training behaviors in the different phases, the term $\tcircle{1}$ is denoted as $\mathcal{L}_{\text{sim}}^{(k)}(\tilde{p}_{*}^{(k)}\Vert p_{*}^{(k)})=-\alpha\tilde{p}_*^{(k)}\text{log}(p_*^{(k)})$ where $\alpha$ is a hyper-parameter to tune its importance. Then, the simple-to-difficulty crime distillation can be defined as:
\begin{small}
\begin{align}
\begin{split}
\mathcal{L}_{_\text{TC}}^{(k)}(\tilde{p}_{*}^{(k)}\Vert p_{*}^{(k)})&= \Gamma( t, \mathcal{L}_{_\text{sim}}^{(k)},\mathcal{L}_{_\text{diff}}^{(k)} )\\&=
\begin{cases}
   \mathcal{L}_{\text{sim}}^{(k)}, & \text{ if } t<\tau_{0},   \\
   \tau_r\mathcal{L}_{\text{sim}}^{(k)} +  (1-\tau_r)\mathcal{L}_{\text{diff}}^{(k)}, & \text{ if } \tau_{0} \le t \le \tau_{1}\\
   \mathcal{L}_{\text{diff}}^{(k)}, & \text{ if } t>\tau_1 \\
\end{cases}
\end{split}
\label{loss_ti}
\end{align}
\end{small}%
where $\tau_0$ and $\tau_1$ ($\tau_0 < \tau_1$) are set to facilitate adequate training across distinct phases. During the intermediate training phase, we incorporate a uniform distribution to generate a random number $\text{rand}(0,1)$ within the range of 0 to 1 in each iteration, thereby enabling a switch in training behaviors. Specifically, $\tau_r$ is assigned a value of 1 when $t<\text{rand}(0,1)$, and 0 otherwise.

\subsection{Less-to-More Non-target Crime Distillation}
\label{sec33}
\paragraph{Analysis about Non-targets.}
The term $\tcircle{3}$ in Eq. (\ref{kd_decouple}) facilitates knowledge transfer among various non-target crime events, often referred to as ``dark knowledge" in literature~\cite{DBLP:conf/icml/FurlanelloLTIA18,Zhao_2022_CVPR}. This knowledge provides insights into the probabilities of unobserved crime intents likely to occur next. Inspired by curriculum learning~\cite{DBLP:conf/icml/BengioLCW09}, we recognize the challenge of directly learning unobserved crime intents among all non-target crime events, since both the quantity and frequency of these events may increase the learning difficulty. To address this challenge, we propose less-to-more non-target crime distillation, constructing a series of smaller sub-tasks by incrementally increasing subsets of crime events based on their frequency.

\paragraph{Non-target Crime Distillation Loss.}
To gradually add non-target crime events into training, a random masking vector $\bm{c}=[...,0,..,1,...]$ is first introduced, where the value of masked crime events is set to 1, or otherwise it is 0. Then, the training progress $t$ is introduced to reduce the number of masked crime events with the evolution of training. Formally, the probability distribution of non-target crime events can be rewritten as:
\begin{small}
\begin{align}
\begin{split}
&\bm{q}_{_\text{NC}}^{(k)}(t)=\left[q_{1}^{(k)},q_{2}^{(k)},...,q_{\left|\mathcal{I}\right|}^{(k)}\right]=\text{softmax}(\bm{z}^{(k)}-1000\cdot\bm{c}), \\
&\text{s.t.}  \quad c_{*}=1  \quad \text{and}  \quad \sum_{i=1,i\neq*}^{\left|\mathcal{I}\right|}c_i=
\begin{cases}
(1-t)\cdot{\left|\mathcal{I}\right|}, & \text{ if } t<\tau_1\\
0, & \text{others}
\end{cases}
\end{split}
\label{logit_ni}
\end{align}
\end{small}%
Here, $c_i$ represents an element in the masking vector $\bm{c}$. $\tau_1$ is set to ensure adequate training of all non-target events whose value is the same with that in target crime distillation in Eq.~(\ref{loss_ti}). To maintain numerical stability, we employ a relatively large constant value of 1000 to nullify the contributions of masked events. Subsequently, the loss for the $k$th peer network is defined as:
\begin{small}
\begin{align}
\begin{split}
\mathcal{L}_{_\text{NC}}^{(k)}\left(\tilde{\bm{q}}_{_\text{NC}}^{(k)}(t) \Vert \bm{q}_{_\text{NC}}^{(k)}(t)\right)
=\text{CE}({\tilde{\bm{q}}_{_\text{NC}}^{(k)}(t)}\|{\bm{q}_{_\text{NC}}^{(k)}(t)})\\
=\beta\cdot\sum_{i=1}^{\left|\mathcal{I}\right|}\tilde{q}_{i}^{(k)}\text{log}(q_{i}^{(k)}).
\end{split}
\label{loss_ni}
\end{align}
\end{small}%

Note that the random masking vector $\bm{c}$ is the major difference of our work. With the advance of training process $t$, the masked non-target crime events will gradually decrease and finally degrade to a one-hot vector, i.e., the label of the next target event $\bm{y}_m$, which makes the loss equal to the term $\tcircle{3}$ in Eq. (\ref{kd_decouple}). In particular, frequency is utilized to sample crime events, where high-frequency events are more likely to occur in different spots and show spot-shared intents.

\subsection{Peer Group Expansion}
\label{sec34}
Our CrimeAlarm works together naturally with various ensemble strategies~\cite{DBLP:conf/aaai/ChenMWF020,DBLP:conf/cvpr/ZhangXHL18} to extend to more peers, in a way that does not contradict the motivation of our distillation strategy. To avoid introducing extra trainable parameters, we follow a simple ensemble approach from~\cite{DBLP:conf/cvpr/ZhangXHL18}. Given $K$ peers, the distillation loss can be defined as follows:
\begin{small}
\begin{align}
\begin{split}
\bar{\mathcal{L}} _{_\text{TC}}^{(k)}= \frac{1}{K-1} \sum^{K}_{j=1,j\neq k} \mathcal{L}_{_\text{TC}}^{(j)}\left(p_{*}^{(j)} \Vert p_{*}^{(k)}\right),\\
 \bar{\mathcal{L}} _{_\text{NC}}^{(k)}= \frac{1}{K-1} \sum^{K}_{j=1,j\neq k} \mathcal{L}_{_\text{NC}}^{(j)}\left(\bm{q}_{_\text{NC}}^{(j)}(t) \Vert \bm{q}_{_\text{NC}}^{(k)}(t)\right),
\end{split}
\label{ext}
\end{align}
\end{small}%
In the above definitions, our proposed CrimeAlarm between two peers can be regarded as a special case of Eq.~(\ref{ext}), where $K=2$. Finally, we jointly optimize label-based loss and the distillation loss as a holistic distillation framework:
\begin{small}
\begin{align}
 \mathop{\min}_{\Theta} \mathcal{L} = \sum_{k=1}^{K} (\mathcal{L}_{_\text{CE}}^{(k)}+\mathcal{L}_{_\text{Crime}}^{(k)})=\sum_{k=1}^{K} (\mathcal{L}_{_\text{CE}}^{(k)}+\bar{\mathcal{L}}_{_\text{TC}}^{(k)}+\bar{\mathcal{L}}_{_\text{NC}}^{(k)}),
\label{joint_learning}
\end{align}
\end{small}%
where $\Theta$ is the parameters of the prediction networks. $\mathcal{L}_{_\text{CE}}^{(k)}=\text{CE}(\bm{y}\|\bm{p}^{(k)})$ is the cross-entropy between the label and the probability distribution of the $k$th peer network. The training for more peers is summarized in Algorithm~\ref{alg_frame}. 

\begin{algorithm}[t]
  \caption{Training of CrimeAlarm (More Peers).}
  \label{alg_frame}
  \begin{algorithmic}[1]
    \Require the training set $(\bm{x}_m,\bm{y}_m)\in \mathcal{D}$, the peer number $K$, the current training epoch $\pi$ and the total epoch $\Pi$, the hyper-parameters $\alpha$, $\gamma$ and $\beta$
    \Ensure the parameters $\Theta$ of CrimeAlarm

	\State Randomly initialize all parameters $\Theta$;
		\For {each training batch $(\bm{x}_m,\bm{y}_m)$}
		    \State Update training progress $t=\pi/\Pi$
		    \State Compute $\bm{p}^{(k)}_{_\text{TC}}$, $\bm{q}^{(k)}_{_\text{NC}}$ in Eqs.~(\ref{p_ti}) and (\ref{logit_ni})
		    \For {each peer encoder $k$}
		    \State Compute $\bar{\mathcal{L}}_{_\text{TC}}^{(k)}$ and $\bar{\mathcal{L}}_{_\text{NC}}^{(k)}$ according to Eq.~(\ref{ext})
		    \State Compute cross-entropy $\mathcal{L}_{_\text{CE}}^{(k)}$ according to the ground-truth
		    \EndFor
		    \State $\mathcal{L} \leftarrow \sum_{k=1}^{K} (\mathcal{L}_{_\text{CE}}^{(k)}+\bar{\mathcal{L}}_{_\text{TC}}^{(k)}+\bar{\mathcal{L}}_{_\text{NC}}^{(k)})$
            \State Update all parameters to minimize $\mathcal{L}$;
    \EndFor

    \State return $\Theta$
  \end{algorithmic}
\end{algorithm}

\section{Experiment}
%This section conducts experiments to answer three questions: (1) What is the performance of CrimeAlarm compared to state-of-the-art methods?; (2)Do different parts in our CrimeAlarm contribute to the prediction performance?; and (3) What is the performance of CrimeAlarm when extending it to different prediction networks and more peers?
% \begin{itemize}
%   \item {\bfseries RQ1:} What is the performance of CrimeAlarm compared to other state-of-the-art methods?
%   \item {\bfseries RQ2:} Do different parts in our CrimeAlarm contribute to the prediction performance?
%   \item {\bfseries RQ3:} What is the performance of CrimeAlarm when extending it to other encoders and more peers?
% \end{itemize}

\begin{table}[t]
\tabcolsep=1.5mm  %设置列宽
\renewcommand\arraystretch{1.05}    %设置行高
  \caption{Dataset statistics.}
	\centering
  \begin{tabular}{cccccccc}
    \toprule
    \multirow{2}{*}{Dataset}&		\multirow{2}{*}{\#Spots}& \multirow{2}{*}{\#Records}&		\multicolumn{4}{c}{Records per spot}& \multirow{2}{*}{\#Event classes}\\
\cline{4-7}
		&					&&		Max.&	Min.&		Avg.&			Std.&			\\
	\midrule
    NYC16& 	3,229& 		473,887&	4,496&	3&		146.76&			429.78&		440	\\
    CHI18& 	2,692&		264,314&	3,525&	3&		98.18&			253.36&	246		\\
  \bottomrule
\end{tabular}
\label{tab:dataset}

\end{table}

\subsection{Experimental Setup}
\paragraph{Dataset.} The police departments from New York City and Chicago provide detailed crime records to describe fine-grained crime events. Specifically, these records contain the precincts, premises, time and date of occurrence, and the crime categories. To perform fine-grained prediction, we divide time into different slots at every 3 hours. A simple event model~\cite{DBLP:journals/ws/HageMSHS11} is employed to describe spots by using geographical fields (precinct and premises), and events by using time slot and event category. The data statistics of New York City in 2016 (NYC16) and Chicago in 2018 (CHI18) are shown in Table~\ref{tab:dataset}. 
Following BERT4Rec~\cite{DBLP:conf/cikm/SunLWPLOJ19}, spots containing fewer than five crime events are excluded. The last event of each spot is reserved for testing, while the event preceding the last serves as the validation set. We set the maximum sequence length to 200. Sequences exceeding this length are divided from right to left. If a sequence falls short of the maximum length, additional ``padding" tokens are appended to the right.

\paragraph{Implementation Details.} 
Our CrimeAlarm~\footnote{https://github.com/hukx-issac/CrimeAlarm} adopts two BERT4Rec~\cite{DBLP:conf/cikm/SunLWPLOJ19} as our default peer prediction networks. It is trained from scratch by using a Adam optimizer (initial learning rate is 1e-3) with linear decay, and a batch size of 256 for 100 training epochs. We configure the model with two multi-head attention layers, each with two attention heads. The embedding dimension is set to 64, while the dimension of the intermediate layer is 256. The thresholds for learning progress, $\tau_0$ and $\tau_1$, are set to 0.2 and 0.7 respectively, ensuring adequate training across different phases. We set the truncation proportion, $\varepsilon$, to 0.01. Hyperparameters $\alpha$, $\gamma$, and $\beta$ are fine-tuned via grid search on the validation set, selecting from values of ${1,3,5,7,9}$, ${0,0.5,1.0,1.5,2.0}$, and ${1,3,5,7,9}$ respectively.

\paragraph{Evaluation Metrics.} 
As multiple crime intents are likely within testing samples, ranking-based metrics are adopted in our evaluation, including top-$N$ Hit Ratio (HR@$N$), Top-$N$ Normalized Discounted Cumulative Gain (NDCG@$N$), and Mean Reciprocal Rank (MRR). $N$ is set as $\{5,10\}$. All metrics are derived from the principle of the higher, the better. To speed up evaluation, each ground-truth is paired with 100 randomly sampled events that the spots have not interacted with, according to their popularity. 

\paragraph{Baselines.}
Our CrimeAlarm is compared against seven state-of-the-art methods, including crime prediction~\cite{hu2021duronet,DBLP:conf/cikm/HuL0LT21,DBLP:journals/tkde/ZhaoLCZ23}, multi-class sequential prediction for different social events~\cite{DBLP:conf/www/FanLWWNZPY22,DBLP:conf/aaai/ZhouZPZLXZ21} and knowledge distillation~\cite{DBLP:conf/cvpr/ZhangXHL18,Zhao_2022_CVPR}.

\begin{itemize}
\item \textbf{DuroNet$^*$}~\cite{hu2021duronet}: A crime event model for predicting crime amounts by using contexts. To adapt for fine-grained event classification, we replace the regression loss with cross-entropy in the original DuroNet model.
\item \textbf{HAIL}~\cite{DBLP:conf/cikm/HuL0LT21}: A fine-grained sequential crime prediction framework that employs mutual exclusivity KD to handle implicitly hard samples.
\item  \textbf{CLC}~\cite{DBLP:journals/tkde/ZhaoLCZ23}: A state-of-the-art crime prediction method using classification-labeled continuousization strategy for sparse classification problem.
 \item \textbf{STOSA}~\cite{DBLP:conf/www/FanLWWNZPY22}: A sequential recommendation method that employs distributions to model dynamic uncertainty over ten thousands of items.
\item \textbf{Informer$^*$}~\cite{DBLP:conf/aaai/ZhouZPZLXZ21}: A general sequence time-series forecasting method that models dynamics by employing long-term electricity consumption context. Similar to Duronet, we replace the objective loss with cross-entropy.
\item \textbf{DKD}~\cite{Zhao_2022_CVPR}: This method distills the knowledge between target and non-target class, separately. Our proposed distillation can degrade to it by eliminating the easy-to-hard learning pipeline. In this paper, we perform DKD based on Transformer~\cite{DBLP:conf/nips/VaswaniSPUJGKP17}.
\item \textbf{DML}~\cite{DBLP:conf/cvpr/ZhangXHL18}: A popular distillation method based on mutual learning. Similarly, we apply DML based on Transformer.
\end{itemize}

To ensure fair comparisons, the same set of settings is applied to the Transformer backbone for our model and the seven baselines. Other hyper-parameters in the baselines are reproduced from their original papers without changes.

\begin{table*}[t]
  \caption{The performance comparison across various methods. The numbers in \textbf{Bold} and \underline{Underline} denote the best result and the strongest baseline, respectively. The row ``Improv.'' means the relative improvement of our best result over the best baseline result. Note that we can keep either peer in the testing phase, since KD makes them produce similar performance results.}
	\resizebox{\linewidth}{!}{
  \begin{tabular}{clccccccc||ccr}
    \toprule
    \multirow{2}{*}{Dataset}&		 \multirow{2}{*}{Metrics}&				  	\multicolumn{3}{c}{Sequential Crime Prediction}&	\multicolumn{2}{c}{Multi-class Prediction}& \multicolumn{2}{c}{Knowledge Distillation}&	    \multicolumn{2}{c}{CrimeAlarm (ours)}& 	\multirow{2}{*}{Improv.}\\
\cmidrule(r){3-5} \cmidrule(r){6-7} \cmidrule(r){8-9}\cmidrule(r){10-11}
& &     DuroNet$^*$&		HAIL& CLC& STOSA& Informer$^*$&		DKD& \multicolumn{1}{c}{DML} & $\text{Peer}~1$&  $\text{Peer}~2$&\\
     \hline
	 \hline
	\multirow{5}{*}{NYC16}	&		NDCG@5&	     0.1777& 0.4131&  \underline{0.4150}& 0.3713& 0.2108& 0.4083& 0.3861& 0.4335& \textbf{0.4337}& 4.51\%\\
						  	&		NDCG@10&     0.2270& \underline{0.4461}&  0.4455& 0.4143& 0.2540& 0.4419& 0.4177& \textbf{0.4642}& 0.4637& 4.06\%\\
						  	&		HR@5&	     0.2926& 0.5070&  \underline{0.5187}& 0.4642& 0.3165& 0.5042& 0.4819& 0.5290& \textbf{0.5302}& 2.22\%\\
						  	&		HR@10&       0.4460& \underline{0.6141}&  0.6125& 0.5974& 0.4500& 0.6079& 0.5785& \textbf{0.6240}& 0.6228& 1.61\%\\
						  	&		MRR&	     0.1833& \underline{0.4113}&  0.4088& 0.3727& 0.2160& 0.4059& 0.3834& \textbf{0.4297}& 0.4295& 4.47\%\\
    \hline
	\multirow{5}{*}{CHI18}	&		NDCG@5&	     0.1757& \underline{0.4684}&  0.4363& 0.4090&0.2200& 0.4683& 0.4501& \textbf{0.5046}& 0.5039& 7.73\%\\
						  	&		NDCG@10&	 0.2236& \underline{0.5026}&  0.4740& 0.4505& 0.2660& 0.5004& 0.4870& \textbf{0.5363}& 0.5361& 6.71\%\\
						  	&		HR@5&	     0.2923& 0.5683&  0.5435& 0.5126& 0.3228& \underline{0.5717}& 0.5457& \textbf{0.6059}& 0.6048& 5.98\%\\
						  	&		HR@10&       0.4421& \underline{0.6742}&  0.6605& 0.6415& 0.4650& 0.6716& 0.6601& 0.7036& \textbf{0.7039}& 4.41\%\\
						  	&		MRR&	     0.1802& \underline{0.4631}&  0.4304& 0.4050& 0.2259& 0.4610& 0.4480& \textbf{0.4969}& 0.4965& 7.30\%\\
    \bottomrule
  \end{tabular}
 }
\label{compbase}
\end{table*}

\subsection{Overall Performance Comparison}
Table~\ref{compbase} presents the performance of the different methods. This leads to the following findings:

\textbf{CrimeAlarm Outperforms All Baselines.} From Table~\ref{compbase}, we find that CrimeAlarm shows better prediction accuracy than all baselines. In particular, it achieves 4.51\% and 7.73\% improvements over the best results in terms of NDCG@5 in the NYC16 and CHI18 datasets, respectively. This observation indicates the superiority of our proposed CrimeAlarm in modeling spot-shared and spot-specific intents by KD in an easy-to-hard pipeline. 

Compared to DuroNet$^*$, STOSA and Informer$^*$, our proposed CrimeAlarm employs soft target to model unobserved criminal intents, and learns spot-shared and spot-specific criminal intents in a meaningful order.
Compared to HAIL and CLC, CrimeAlarm decouples the probability distributions into target and non-target parts. This reduces the suppression of non-target probabilities, and balances the learning of their knowledge. Compared to DKD and DML, we further incorporate curriculum learning into KD, enhancing distillation over numerous candidate crime events. 

\textbf{Small Performance Difference Between Peers.} CrimeAlarm employs multiple peers, each of which generates a prediction result. Deciding which prediction to use in the testing stage presents a challenge. However, we discovered that the performance difference between Peer 1 and Peer 2 is negligible, as KD allows them to learn from each other. This means that the selection of encoders can be random.

\subsection{Ablation Study}
\label{Ablation}
We set four variants of our CrimeAlarm as follows:
\begin{itemize}
  \item $\neg$CTC: This variant removes the difficult target crime distillation in the later training stage, i.e., $\mathcal{L}_{\text{diff}}^{(k)}$ in Eq.~(\ref{loss_ti}), and keeps $\mathcal{L}_{\text{sim}}^{(k)}$.
  \item $\neg$TC: This variant removes the whole target crime distillation, i.e., $\mathcal{L} = \sum_{k=1}^{K} (\mathcal{L}_{_\text{CE}}^{(k)}+\mathcal{L}_{_\text{NC}}^{(k)})$
  \item $\neg$CNC: This variant removes the less-to-more non-target crime distillation in Eq.~(\ref{logit_ni}) and just adopts the conventional manner in Eq.~(\ref{kd_decouple}).
  \item $\neg$NC: This variant removes the whole non-target crime distillation, i.e., $\mathcal{L} = \sum_{k=1}^{K} (\mathcal{L}_{_\text{CE}}^{(k)} +\mathcal{L}_{_\text{TC}}^{(k)})$.
\end{itemize}

\begin{figure}[t]  
\centering 
  \begin{minipage}[t]{0.66\linewidth} % 如果一行放2个图，用0.5，如果3个图，用0.33  
    \centering   
    \includegraphics[width=\linewidth]{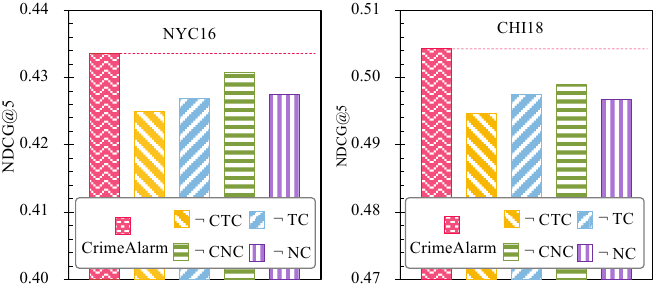}
  \caption{Ablation study.}
  \label{fig:ablation}
  \end{minipage}%   
  \begin{minipage}[t]{0.3\linewidth}   
    \centering   
    \includegraphics[width=0.85\linewidth]{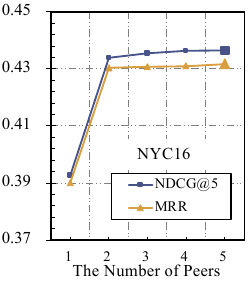}  
    \caption{Growing peers.}
    \label{fig:multi-peer}
  \end{minipage}   
\end{figure} 

The average results of two peers in terms of NDCG@5 are reported in Fig.~\ref{fig:ablation}. ``$\neg$'' indicates the removal of the corresponding distillation in the training phase, while the remainders are kept. We have the following observations:

\textbf{Effectiveness of Simple-to-Difficult Target Crime Distillation.}
Observing the variant $\neg$CTC (yellow bar), we observe a decline in performance following the removal of difficult target crime distillation. Furthermore, comparing $\neg$CTC and $\neg$TC (blue bar) shows a negative impact of $\mathcal{L}_{\text{sim}}^{(k)}$ in vanilla knowledge distillation. These findings align with prior research~\cite{Zhao_2022_CVPR}, indicating the effectiveness of our target crime distillation in enhancing performance, particularly by emphasizing spot-specific intents in low-confidence sequences.

\textbf{Effectiveness of Less-to-More Non-target Crime Distillation.}
Comparison against the variant $\neg$CNC (green bar) consistently shows CrimeAlarm's superior performance. Furthermore, upon removing the entire non-target crime distillation, $\neg$NC (purple bar) experiences a performance decline. This observation underscores the significance of knowledge transfer among various non-target crime events in capturing dynamic intents. Direct exposure of global knowledge among non-target events may not yield optimal results.

\subsection{Further Probing}
\subsubsection{Distillation with More Peers.}
We follow the ensemble manner in~\cite{DBLP:conf/cvpr/ZhangXHL18}, and report the results of extended CrimeAlarm in Fig.~\ref{fig:multi-peer}. We find that the performance slowly rises with the increment of peers, indicating there is an upper limit of peers. Note that it is inevitable that computation cost in the training phase will increase, since more peers bring more parameters and gradient information.

\begin{table}[t]
\tabcolsep=1.8mm  %设置列宽
\renewcommand\arraystretch{1.05}    %设置行高
\centering
\caption{Performance improvement when combining different sequence encoders.}
\resizebox{0.7\linewidth}{!}{
\begin{tabular}{cccccccc}
\toprule
\multirow{2}{*}{Structure} & \multicolumn{1}{c}{\multirow{2}{*}{Distillation}}    & \multicolumn{3}{c}{NYC16}   & \multicolumn{3}{c}{CHI18}\\
\cmidrule(r){3-5}\cmidrule(r){6-8}
\multicolumn{1}{c}{}     && NDCG5   & HR@5    & MRR     & NDCG5   & HR@5   & MRR    \\
\midrule
\multirow{2}{*}{GRU} & $\times$         & 0.3549  & 0.4528 & 0.3530&  0.4467& 0.5475& 0.4422   \\
&\checkmark    & 0.4040  & 0.5009 & 0.4010&    0.4607& 0.5615& 0.4552\\
\midrule
\multirow{2}{*}{TCN} & $\times$         & 0.3093  & 0.4007 & 0.3170&  0.2965& 0.3770& 0.3093\\
&\checkmark                             & 0.3278  & 0.4165 & 0.3294& 0.3180& 0.4101& 0.3244\\
\midrule
\multirow{2}{*}{Transformer} & $\times$       & 0.3927  & 0.4908 & 0.3901& 0.4619& 0.5638& 0.4562\\
& \checkmark          & 0.4336  & 0.5296 & 0.4296& 0.5043& 0.6054& 0.4967\\
\bottomrule
\end{tabular}
}
\label{tab:diff_encoder}
\end{table}

\subsubsection{Integration with Different Sequence Encoders.}
\label{integration_diff_encoders}
We further attempt other representative prediction methods with different sequence structures, i.e., GRU~\cite{DBLP:journals/corr/HidasiKBT15}) and TCN~\cite{DBLP:journals/corr/abs-1803-01271}. In Table~\ref{tab:diff_encoder}, we can observe that CrimeAlarm enables significant performance improvements. Furthermore, GRU with distillation even shows superior performance over Transformer without distillation in NYC16. This observation demonstrates the effectiveness and flexibility of our CrimeAlarm model.

\begin{figure}[t]  
\centering 
  \begin{minipage}[t]{0.38\linewidth} 
    \centering   
    \includegraphics[width=0.85\linewidth]{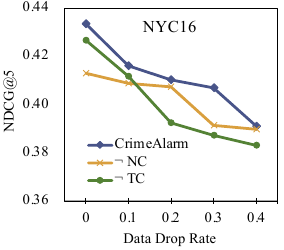}
  \caption{Discussion on densities.}
  \label{fig:density}
  \end{minipage}%   
  \begin{minipage}[t]{0.62\linewidth}   
    \centering   
    \includegraphics[width=\linewidth]{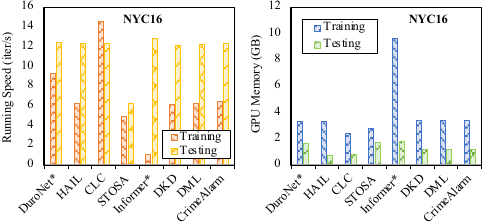}
\caption{Efficiency comparison.}
 \label{fig:efficiency}
  \end{minipage}   
\end{figure}

\subsubsection{Effect of Crime Data Density.}
\label{density}
As crime data present relatively large density differences, we further investigate the effect of data densities on NYC16. Specifically, we randomly drop some crime records within an event sequence according to a certain drop rate. The length of the new sequence can be computed as: $\text{seq\_new = seq\_old * (1-drop\_rate)}$. The performance variation is in Fig.~\ref{fig:density}.

We have two main observations: (1) The performance of CrimeAlarm is consistently superior over its variants in terms of different data drop rate; and (2) The performance of variant $\neg$TC first outperforms the variant $\neg$NC, and then falls quickly with the increment of data drop rate. The above observations indicate that a long sequence (high density) may be easy to express multiple crime intents. And, non-target crime distillation can model unobserved intents. When the data density is low, a short sequence may contain less crime intents. The proposed target crime distillation is sufficient to learn observed intents well by following a simple-to-difficult pipeline. This also proves the different effects of target and non-target crime distillation.   

\subsubsection{Efficiency}
\label{Efficiency}
All methods were implemented on a platform environment comprising TITAN XP GPU with 12G memory, Python 3.8 and Pytorch 1.10. The running speed and memory cost are reported in Fig.~\ref{fig:efficiency}.

As the learning of multiple intents or distribution will result in more parameter and gradient computation, such methods present relatively low iteration speeds and high memory compared to DuroNet in the training phase. Our proposed CrimeAlarm achieves comparable iteration speed, compared to other baselines (e.g., HAIL, STOSA, DKD and DML). Furthermore, by retaining only one peer prediction network during the testing phase, CrimeAlarm's efficiency can be significantly improved, achieving speeds and memory usage comparable to baselines. This indicates that CrimeAlarm does not incur additional costs during the testing phase or model deployment.

\begin{figure}[t]
  \centering
    	\includegraphics[width=0.9\linewidth]{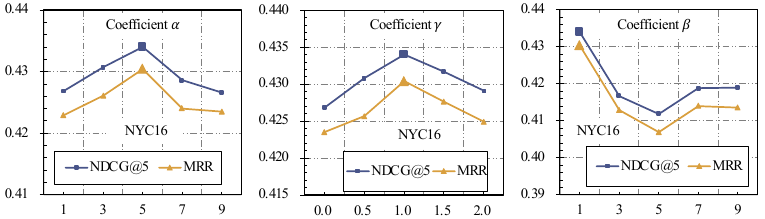}
    \caption{Parameter Sensitivity on NYC16. Other hyper-parameters are unchanged.}
  \label{fig:para_sens}
\end{figure}

\subsubsection{Parameter Sensitivity}
After the grid search, the final hyper-parameter groups of CrimeAlarm $(\alpha,\gamma,\beta)$ are $(5,1,1)$ and $(3,1,3)$ for NYC16 and CHI18, respectively. We further investigated performance under optimal settings, and present the averages of metrics on the testing datasets in Fig.~\ref{fig:para_sens}.

When increasing $\alpha$ and $\gamma$, the performance first rises and then falls after a certain threshold. It indicates overemphasizing either of them will overemphasize high- or low-confidence sequences, leading to ignorance of shared or specific intents. With few exceptions, we find performance degrades with the increment of $\beta$, showing under-fitting in non-target crime distillation.

\section{Related Work}
\paragraph{\textbf{Sequential Crime Prediction.}}
State-of-the-art sequential crime prediction has already been developed from statistical methods~\cite{DBLP:conf/smartcomp/CatlettCTV18} to deep learning. According to prediction granularity, deep learning methods can be summarized into two groups. (1) \textbf{Crime amounts}: This group~\cite{hu2021duronet,DBLP:conf/ijcai/Yi0ZG19,DBLP:conf/icdm/Yi0ZZX18,DBLP:conf/aaai/ZhaoFLT22} just predicts crime amounts in the near future. They generally model spatio-temporal correlations to capture dynamic criminal intents. Based on crime data, some works further introduce multi-modal data to enrich spatio information, such as 311 complaint data~\cite{DBLP:conf/icdm/ZhaoT17}, and meteorology~\cite{DBLP:conf/cikm/ZhaoT17}. (2) \textbf{Crime category}: The second group predicts the specific crime category of each urban region by analyzing criminal intents across categorical dependencies. In particular, most works~\cite{DBLP:conf/www/HuangZZWCY19,DBLP:conf/cikm/HuangZZC18,DBLP:conf/aaai/WangLYSYS22} model the sequential intents of each crime category separately and predicate the binary crime occurrence of all crime categories simultaneously. More recently, HAIL~\cite{DBLP:conf/cikm/HuL0LT21} focuses on observed target intents that are hard to learn and ranks crime categories via probability distribution. CLC~\cite{DBLP:journals/tkde/ZhaoLCZ23} proposes a label preprocessing strategy to support a smaller temporal
granularity crime prediction. 

The above crime prediction methods do not fully model intent dynamics and capture unobserved intents. Moreover, some multi-class sequential prediction methods in other scenarios addressed similar problems of intent dynamics. These methods are also related to our work and compared in our experiments, such as web recommendation~\cite{DBLP:conf/www/FanLWWNZPY22,DBLP:conf/cikm/SunLWPLOJ19,DBLP:conf/dasfaa/TianHZWLS21} and electricity consumption~\cite{DBLP:conf/aaai/ZhouZPZLXZ21}.

\paragraph{\textbf{Knowledge Distillation.}}
KD is an effective approach to transfer knowledge from the teacher model to student model. Within the scope of online mutual distillation~\cite{DBLP:conf/cvpr/ZhangXHL18}, our CrimeAlarm involves with logit knowledge. Lots of related studies~\cite{DBLP:conf/iccv/ChoH19,DBLP:conf/aaai/MirzadehFLLMG20} exert regularization~\cite{DBLP:journals/pami/WangY22} on soft label. Recent DKD~\cite{Zhao_2022_CVPR} redefines the decoupled formulation of logit distillation. Our work is different from these studies by incorporating curriculum learning into KD.

\section{Conclusion and Future Work}
This work highlighted intensive intent dynamics in fine-grained sequential crime prediction. We proposed a novel prediction framework, CrimeAlarm, to capture spot-shared and spot-specific intents by following an easy-to-hard pipeline. With extensive experiments on the real-world crime datasets, our results indicated that CrimeAlarm achieves superior performance over baseline models. For future work, we will delve into spatial modeling and expand our CrimeAlarm framework to more scenarios, such as urban anomalies and web behaviors.

\subsubsection{Acknowledgement}
This work was supported in part by the National Natural Science Foundation of China (62276196,62072257), the Australian Research Council (DP22010371,LE220100078), the Key Research and Development Program of Hubei Province (2021BAA030) and the China Scholarship Council (202106950041).

\bibliographystyle{splncs04}
\bibliography{CrimeAlarm}




% LNCS Homepage, \url{http://www.springer.com/lncs}. Last accessed 4
% \end{thebibliography}
\end{document}